\documentclass[conference,a4paper]{IEEEtran}
\IEEEoverridecommandlockouts

\usepackage[hidelinks]{hyperref}
\usepackage[cmex10]{amsmath}
\usepackage{amssymb,amsfonts}
\usepackage{dblfloatfix}

\usepackage[ruled,vlined]{algorithm2e}
\usepackage{graphicx}
\graphicspath{{Figures/PDF/}{Figures/PNG/}}

\usepackage{booktabs}
\usepackage{siunitx}
\usepackage[numbers,compress]{natbib}
\usepackage{texnames}
\usepackage{bm,bbm}
\usepackage{orcidlink}
\usepackage[table]{xcolor}

\newcommand{\Ihr}{I_{\text{HR}}}
\newcommand{\Hhr}{H_{\text{HR}}}
\newcommand{\Whr}{W_{\text{HR}}}
\newcommand{\Imr}{I_{\text{MR}}}
\newcommand{\Hmr}{H_{\text{MR}}}
\newcommand{\Wmr}{W_{\text{MR}}}

\usepackage{tikz}

\newcommand\copyrighttext{%
  \footnotesize \textcopyright 2025 IEEE. Personal use of this material is permitted.
  Permission from IEEE must be obtained for all other uses, in any current or future
  media, including reprinting/republishing this material for advertising or promotional
  purposes, creating new collective works, for resale or redistribution to servers or
  lists, or reuse of any copyrighted component of this work in other works.}
\newcommand\copyrightnotice{%
\begin{tikzpicture}[remember picture,overlay]
\node[anchor=south,yshift=10pt] at (current page.south) 
  {\fbox{\parbox{\dimexpr\textwidth-\fboxsep-\fboxrule\relax}{\copyrighttext}}};
\end{tikzpicture}%
}

\begin{document}

\title{Cross-Scale Pretraining: Enhancing Self-Supervised Learning for Low-Resolution Satellite Imagery for Semantic Segmentation
}

\author{	\IEEEauthorblockN{John Waithaka\orcidlink{0009-0003-1574-888X} }
	\IEEEauthorblockA{\textit{Carnegie Mellon University Africa}\\
		Kigali, Rwanda\\
		jwaithak@andrew.cmu.edu}
	\and
	\IEEEauthorblockN{Gustave Bwirayesu\orcidlink{0009-0002-9546-2293}}
	\IEEEauthorblockA{\textit{Carnegie Mellon University Africa}\\
		Kigali, Rwanda\\
		gbwiraye@andrew.cmu.edu}
	\and
	\IEEEauthorblockN{Moise Busogi\orcidlink{0000-0002-5245-0113}}
	\IEEEauthorblockA{\textit{Carnegie Mellon University Africa}\\
		Kigali, Rwanda\\
		mbusogi@andrew.cmu.edu}
}

\maketitle
\copyrightnotice

\begin{abstract}
    Self-supervised pretraining in remote sensing is mostly done using mid-spatial resolution (MR) image datasets due to their high availability. Given the release of high-resolution (HR) datasets, we ask how HR datasets can be included in self-supervised pretraining to enhance MR image representation learning and downstream segmentation performance on MR tasks. We design a spatial affinity component that can be added to existing self-supervised learning frameworks and that uses HR imagery to learn better representations of MR imagery. We test the spatial affinity component on two self-supervised learning frameworks and show that it outperforms models pretrained on HR or MR images alone.
\end{abstract}


\section{Introduction}

Semantic segmentation is an important task in remote sensing, enabling, for example, the extraction of crop cover, flood extent and marine pollution maps from satellite imagery for applications in food security, disaster management, and climate research. While deep learning has excelled in this task, its performance is limited by the scarcity of pixel-level annotations in Earth Observation, which are costly to acquire. Pretraining is commonly used to improve task performance in such annotation-scarce settings. Self-Supervised Learning (SSL) is a particularly fitting pretraining approach for the remote sensing field, where annotations are scarce but imagery data are abundant given SSL does not require annotation \cite{wang2023ssl4eo, cong2022satmae}. 

Unlike general computer vision, remote sensing has multi-sensor multi-spatial resolution imagery data. Higher resolution (HR) datasets naturally yield superior performance on tasks like semantic segmentation and object detection due to their richer spatial detail \cite{Shermeyer2019TheImageryimagery}. However, HR imagery is often costly, proprietary, or subject to access restrictions. On the other hand, mid-resolution (MR) datasets are much more accessible but yield suboptimal task performance. Most remote sensing SSL literature has focused on MR imagery (i.e. $\geq$10m ground sample distance (GSD)) datasets such as Sentinel 1/2 for pretraining \cite{cong2022satmae, reed2023scaleMAE, nedungadi2024mmearth, szwarcman2025prithvi, tseng2025galileo, brown2025alphaearth, jakubik2025terramind}.
We argue that this is because of the high availability of massive and diverse MR datasets suitable for self-supervised pretraining, rather than any inherent performance advantage.
Recently, large and diverse historical high-resolution (i.e. $\leq5$m GSD) and mid-resolution image pair datasets such as Sen2Venus \cite{sen2venus}
have been made public, allowing pretraining with HR data. A straightforward approach would be to pretrain on the HR data alone as in DINOv3 \cite{simeoni2025dinov3}. 
However, in following this approach, we lose the benefits of mid-resolution satellite imagery such as the higher spectral dimension of most mid-resolution satellite imagery, which has been shown to improve transfer performance \cite{cong2022satmae}. 
Further, most downstream tasks will use mid-resolution imagery for input due to its high availability. Intuitively, a pretrained model that has not encountered mid-resolution imagery during pretraining may not transfer as well on mid-resolution downstream datasets as one that has.

Therefore, in order to exploit the richer visual detail of HR satellite images without losing the benefits of MR images, we propose a cross-scale pretraining strategy using a real high- and mid-resolution pair dataset; specifically, we use the high-resolution images to learn richer representations of the mid-resolution images. To achieve this, we design the simple \textit{spatial affinity component} that can be added to existing SSL schemes. This component's purpose is to learn mid-resolution image patch representations that contain the level of spatial detail present in high-resolution images.

We compare models pretrained with the spatial affinity component using HR/MR pair imagery to models pretrained solely with HR or MR imagery on four semantic segmentation MR tasks. We find that adding the spatial affinity component and training with real HR and MR image pair data outperforms pretraining solely with MR or HR imagery.

\section{Related Work}
\subsection{Self-Supervised Learning for Remote Sensing}
Self-supervised learning for remote sensing has gained significant attention in recent years. Since there are massive remote sensing imagery datasets, only very few of which are annotated, SSL naturally fits the remote sensing domain. SSL is used to pretrain foundation models, which are then tuned for specific downstream applications. Most earlier works in SSL for remote sensing use contrastive learning strategies \cite{ Hou2022HyperspectralLearning, Ayush2021Geography-AwareLearning, Jean2019Tile2Vec:Data, Wang2024Multilabel-GuidedPretraining}, which learn by maximising similarity between the embeddings of related samples and minimising similarity of unrelated samples. Recently, however, focus has shifted to masked image modelling (MIM) \cite{Gao2022AClassification, cong2022satmae, reed2023scaleMAE, Tang2023Cross-ScaleSensing, nedungadi2024mmearth, tseng2025galileo, szwarcman2025prithvi}. MIM strategies learn by reconstructing masked pixels or patch representations \cite{he2022MAE, wei2024latentMIM}. The state-of-the-art remote sensing SSL frameworks are MIM models.

Most prior remote sensing SSL works pretrain with MR satellite image datasets. We argue that this is because mid-resolution datasets (e.g. Sentinel 1/2) are much more available than higher resolution datasets, and not because this is optimal. We compare the downstream performance of models pretrained with HR data only, MR data only, and both HR and MR data.




\subsection{Multi-resolution SSL for Remote Sensing}
Among the various SSL algorithms for remote sensing, some include a component of super-resolution to enable generalisation across multiple resolutions of satellite imagery. These are Scale-MAE \cite{reed2023scaleMAE} and Cross-Scale MAE \cite{Tang2023Cross-ScaleSensing}. These schemes mathematically downsample high-resolution imagery \cite{Christie2018FunctionalWorld} to get the lower resolution counterpart, which is then included in solving the schemes' pretext tasks.  However, research on super-resolution \cite{Chen2019CameraSuper-resolution, Cai2019TowardModel} show that downsampling high-resolution images to get a low-resolution counterpart is much less effective than using a real low-resolution counterpart since downsampling cannot reproduce the physical and sensor-specific characteristics of a real low-resolution image. We pretrain on datasets with real low- and high-resolution pairs.

\subsection{Latent-Space Super-Resolution}
Much of the research in super-resolution focus on super-resolution in pixel-space. Often, the end goal of these works is a higher resolution image. Our work is inspired by Perception-GAN (PGAN) \cite{Li2017PerceptualDetection} which does the super-resolution in latent-space. This involves getting the latent representations for both high- and low-resolution images then minimizing the error between them. PGAN and other works that do super-resolution in latent-space primarily use super-resolution as a means of improving performance on another end goal, such as vehicle detection in satellite imagery \cite{Li2021Target-GuidedImages} and general small object detection \cite{Li2017PerceptualDetection}. Our spatial affinity component can be viewed as a latent-space super-resolution component - a means of learning to represent low-resolution satellite images with the level of spatial and visual detail only contained in higher resolution satellite images.

\subsection{LatentMIM and I-JEPA}
To test whether our spatial affinity component generalises across different SSL algorithms, we evaluate it on two SSL schemes - LatentMIM \cite{wei2024latentMIM} and I-JEPA \cite{assran2023IJEPA}. Both are masked image modelling schemes that reconstruct patch representations rather than pixels. They use a sampled set of image patches (visible patches) to predict the latent representations of the other patches (masked patches). I-JEPA samples the masked patches as blocks and the visible patches block thus becomes the complement of these. LatentMIM samples a random set of non-contiguous patches, $\sim10\%$ of an image's patches, and the rest become the masked patches. The decoder, which is responsible for the reconstruction, takes the visible patch representations and \textit{mask tokens} as input in both settings. Mask tokens are placeholders for the masked patch representations. The I-JEPA decoder passes the two sets of inputs through several self-attention layers and outputs the processed mask tokens as the predicted mask patch representations. The LatentMIM decoder, on the other hand, passes the two sets of inputs through cross-attention layers such that the mask tokens are processed conditioned on the visible patch representations.

\section{Methodology}\label{sec:method}



\subsection{Spatial Affinity Component} \label{subsec:sa-component}
\begin{figure}
    \centering
    \includegraphics[width=0.75\linewidth]{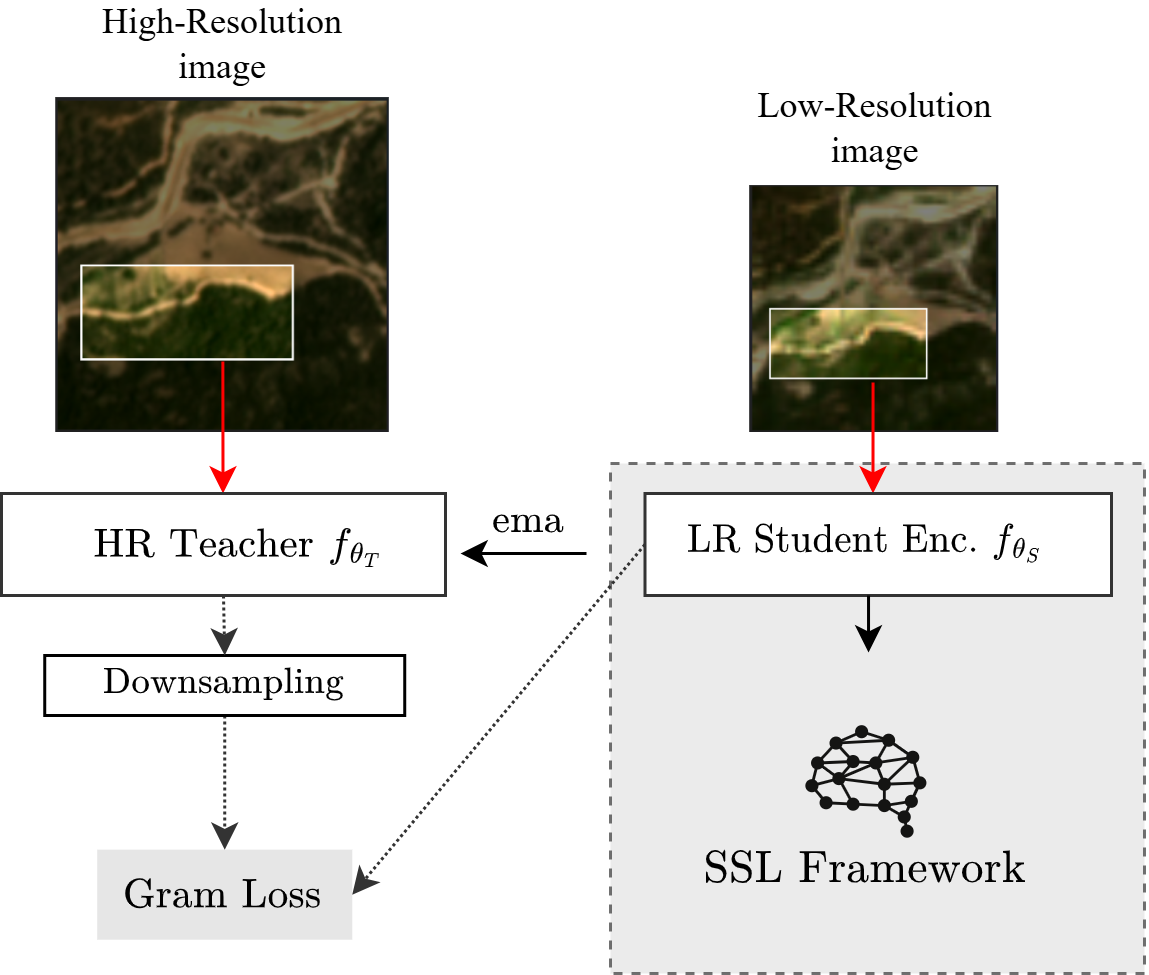}
    \caption{\textbf{Spatial affinity component} samples patches from the high- and mid-resolution inputs. It uses the SSL framework's encoder to encode the lower resolution image and an added high-resolution teacher to encode the high-resolution input. The resulting representations from either encoder are used to compute the gram loss.}
    \label{fig:sa-comp}
\end{figure}

In order to learn mid-resolution image patch representations with the level of spatial detail approaching that of counterpart high-resolution images, we design a spatial affinity component. The spatial affinity component is designed to be added to existing SSL frameworks (see Fig. \ref{fig:sa-comp}). It has a student-teacher architecture for self-distillation \cite{Caron2021DINO}.
The student encoder takes patches of the mid-resolution image as input and is updated through backpropagation during training. The student encoder is shared by the SSL scheme and the spatial affinity component. The teacher encoder takes patches of the high-resolution image as input, and we thus call it the \textit{high-resolution teacher}. This high-resolution teacher has the same architecture as the student encoder and its parameters are updated through an exponential moving average of the student encoder's parameters to prevent collapse \cite{Caron2021DINO}. 

\paragraph{Inputs} To maintain the image size difference between the mid-resolution image $\Imr$ (of size $\Hmr \times \Wmr$) and its high-resolution counterpart $\Ihr$ (of size $\Hhr \times \Whr$), we use a scale factor of $s$ such that \begin{equation}
    \frac{\Hhr}{\Hmr} = \frac{\Whr}{\Wmr} = s \label{eq:scale}
\end{equation}
We use $s = 2$.

As with standard Vision Transformers (ViTs) \cite{dosovitskiy2020ViT}, 
both images are divided into non-overlapping $P \times P$ patches. The input to the student encoder is a set of patches sampled from the mid-resolution image, $\Imr$. We use the default sampling strategies that the host SSL frameworks use---block sampling for I-JEPA  and random sampling for LatentMIM. We also perform ablations on block sampling for LatentMIM. 


Due to the size difference between the image pairs, each patch in $\Imr$ corresponds to $s^2$ patches in $\Ihr$. If $(u, v)$ is the 2D coordinate of a patch in $\Imr$, the corresponding set of patches in $\Ihr$ would have coordinates $\{ (s \cdot u + i, s \cdot v + j)\; |\; 0 \leq i, j < s\}$. After encoding, the set of $s^2$ patch representations corresponding to a patch in $\Imr$ is downsampled to one representation, to match the size of the student encoder output. We perform ablations on bilinear, bicubic and linear projection downsampling methods.


\paragraph{Gram loss} We use gram loss as introduced and defined in DINOv3 \cite{simeoni2025dinov3}. 
This is the mean squared error of the gram matrices of the student encoder output and the downsampled high-resolution teacher output. The gram matrix is a pairwise dot product of $\mathbf{L}_2$-normalised patch representations. Let $Z_S \in \mathbb{R}^{N \times d}$ (respectively $Z_T \in \mathbb{R}^{N \times d}$) be the $\mathbf{L}_2$-normalised patch representations of the student encoder (respectively the high-resolution teacher, after downsampling), then the gram loss is
\[ \mathcal{L} = \|Z_S \cdot Z_S^\top - Z_T \cdot Z_T^\top \|_2^2 \]
$N$ is the number of patch representations and $d$ is the size of each representation.

Using the gram matrices rather than the patch representations gives tolerance for sensor-specific differences in the high- and mid-resolution image pairs while penalising differences in patch-level spatial structure. Different from DINOv3's ``gram teacher" \cite{simeoni2025dinov3} 
whose parameters are selected from an earlier training iteration, our high-resolution encoder is updated along with the student encoder, as our goal in using the gram loss is to tolerate sensor-specific differences rather than maintain patch-level consistency across training progression.

\subsection{Implementation and Evaluation}
We pretrain both SSL schemes on the Sen2Venus dataset \cite{sen2venus}, which contains Sentinel 2 10-metre spatial resolution and Venus 5-meter spatial resolution image pairs collected on the same day.  We sample a random 119,659 image pairs and use only the red, green, blue and near-infrared bands. We train I-JEPA with a learning rate of 0.001 (cosine schedule), batch size of 64 and weight decay of 0.04 with AdamW, and LatentMIM with a learning rate of 0.00015, a batch size of 128, and weight decay of 0.05 with AdamW. Both frameworks are pretrained for 300 epochs.

For I-JEPA \cite{assran2023IJEPA}, we use the ViT-Small architecture with a patch size of 14 for the target encoder, context encoders and high-resolution teacher, and a depth of 12 and embedding size of 384 for the predictor. For LatentMIM, we use the ViT-Small architecture with a patch size of 16 for the online encoder, target encoder and high-resolution teacher. We maintain LatentMIM's decoder depth of 3.

We evaluate via linear probing on a diverse set of earth observation segmentation tasks. These are Geo-Bench's m-SA-Crop-Type \cite{Lacoste2023GEO-Bench:Monitoring}, Sen1Floods11 flood mapping \cite{Bonafilia2020Sen1Floods11:Sentinel-1}, PASTIS crop mapping \cite{Garnot2021PanopticNetworks}, and MADOS marine pollutants and surface features detection \cite{Kikaki2024DetectingImagery}. We use the red, green, blue and near-infrared bands and report mean Intersection over Union (mIoU) averaged over three runs.


\section{Results}
To demonstrate the superior semantic segmentation performance of using the spatial affinity component, we compare three categories of models:
\begin{enumerate}
    \item \textit{MR-model}, pretrained on mid-resolution Sentinel-2 images only,
    \item \textit{HR-model}, pretrained on high-resolution Venus images only, and
    \item \textit{SA-model}, pretrained with the Spatial Affinity component on both mid- and high-resolution images.
\end{enumerate}

The \textit{MR-model} and \textit{HR-model} are pretrained with the same input image size. Further, the mid-resolution input size of the \textit{SA-model} matches the input size of the MR- and HR models.

Table \ref{tab:main-res} shows the downstream segmentation performance of the three models pretrained with I-JEPA and LatentMIM frameworks. We see that the \textit{SA-model} out-performs the LR-model in all but one case. Further, \textit{SA-model} outperforms the HR-model in most cases.



\newcommand{\std}[1]{$\pm$\text{\scriptsize #1}}
\newcommand{\res}[2]{$#1$ $\pm$\text{\scriptsize $#2$}}
\setlength{\tabcolsep}{3pt} 
\begin{table}[hbt]
    \centering
    \caption{Linear probing semantic segmentation performance comparison across SSL frameworks and segmentation tasks}\label{tab:main-res}
    \begin{tabular}{l *{4}{c}}
        \toprule
        & \multicolumn{4}{c}{\textbf{mIoU}} \\ \cmidrule(lr){2-5}
        \textbf{Model} & {\textbf{m-SA}} & {\textbf{Sen1Floods11}} & {\textbf{PASTIS}} & {\textbf{MADOS}} \\ 
        \midrule
        \multicolumn{4}{l}{\textit{LatentMIM Framework}} \\ 
        \midrule
        MR-model & \res{21.41}{0.21}  & \res{76.83}{0.09} & \res{19.60}{0.24} & \res{41.49}{0.12} \\
        
        HR-model & \res{22.39}{0.17} & \res{75.48}{0.05} & \res{18.20}{0.13} & \res{\mathbf{42.94}}{0.14} \\
        
        \rowcolor{gray!30} SA-model & \res{\mathbf{23.39}}{0.02} & \res{\mathbf{78.82}}{0.12} & \res{\mathbf{22.23}}{0.04} & \res{\mathbf{43.00}}{0.84} \\

        \midrule

        \multicolumn{4}{l}{\textit{I-JEPA Framework}} \\ 
        \midrule
        MR-model & \res{20.49}{0.00} & \res{76.34}{0.00} & \res{13.39}{0.00} & \res{36.85}{0.01} \\
        
        HR-model & \res{21.33}{0.00} & \res{76.86}{0.00} & \res{\mathbf{16.34}}{0.00} & \res{\mathbf{42.33}}{0.01} \\
        
        \rowcolor{gray!30}
        SA-model & \res{\mathbf{22.36}}{0.00} & \res{\mathbf{79.04}}{0.00} & \res{15.88}{0.00} & \res{31.79}{0.00} \\

        \bottomrule
    \end{tabular}
\end{table}

\subsection{Super-Resolution}
We investigate whether the downstream improvements are from the informational content of real high-resolution data or are merely a result of the addition of the spatial affinity objective. To test this, we compare the \textit{SA-model} against a control model trained with `false' high-resolution images generated via interpolation of the mid-resolution images. As shown in Table \ref{tab:sr-res}, the model that uses real HR data outperforms the upsampled baseline in all but one case, confirming the value of the high-resolution data in learning better representations of low-resolution data.


\begin{table}[hbt]
	\centering
	\caption{Impact of real vs. false high-resolution imagery on downstream semantic segmentation performance}\label{tab:sr-res}
	\begin{tabular}{l *{4}{S[table-format=2.2]}}
		\toprule
		& \multicolumn{4}{c}{\textbf{mIoU}} \\ \cmidrule(lr){2-5}
		\textbf{Model} & {\textbf{m-SA}} & {\textbf{Sen1Floods11}} & {\textbf{PASTIS}} & {\textbf{MADOS}} \\ 
		\midrule
		\multicolumn{4}{l}{\textit{I-JEPA Framework}} \\ 
		\midrule
		\rowcolor{gray!30} w/ real HR & $\mathbf{23.39}$ & $\mathbf{78.82}$ & $\mathbf{22.23}$ & $\mathbf{43.00}$ \\
		w/ false HR                   & $\mathbf{23.45}$ & 77.69 & 20.94 & 41.17 \\
		\midrule
		\multicolumn{4}{l}{\textit{LatentMIM Framework}} \\ 
		\midrule
		\rowcolor{gray!30} w/ real HR & $\mathbf{22.36}$ & $\mathbf{79.04}$ & $\mathbf{15.88}$ & $\mathbf{31.79}$ \\
		w/ false HR                   & 21.06 & 73.22 & 12.14 & 31.28 \\
		\bottomrule
	\end{tabular}
\end{table}

\subsection{Qualitative View}
\begin{figure}[hbt]
    \centering
    \includegraphics[width=\linewidth]{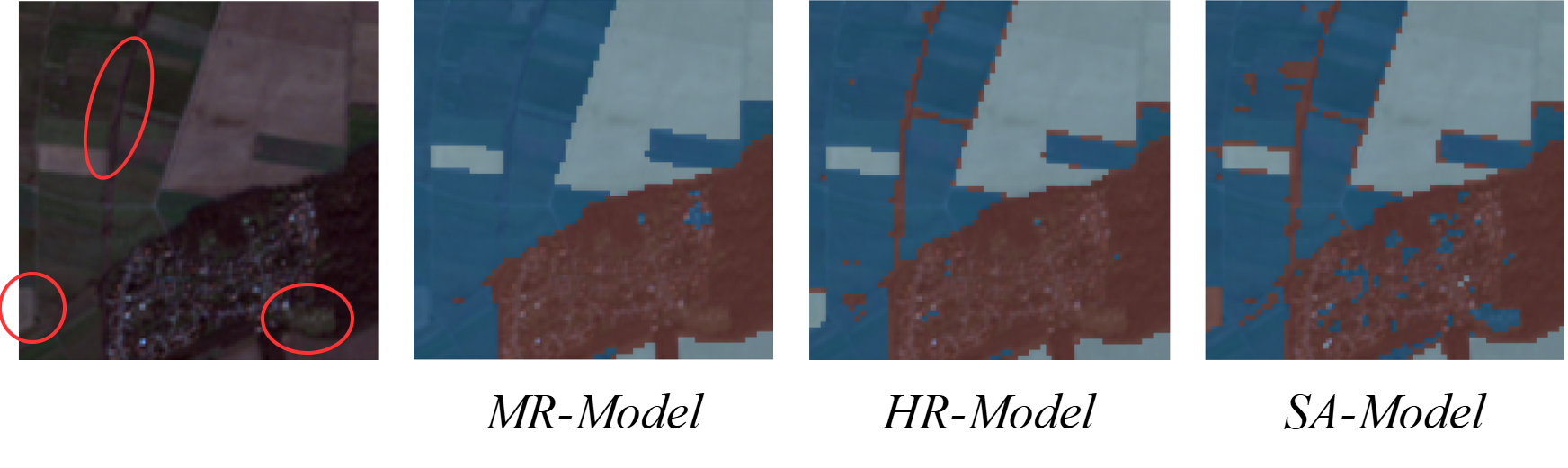}
    \caption{Unsupervised cluster maps of the patch representations of a Sentinel 2 image with $k=3$. Zoom in to see which model's representations are able to identify the distinct features circled in red.}
    \label{fig:cluster-maps}
\end{figure}

We use the \textit{MR-} and \textit{HR-} and \textit{SA-models} pretrained with I-JEPA to generate $64 \times 64$ patch representations of a Sentinel 2 image and cluster them using unsupervised hierarchical clustering into three classes to produce the cluster maps in Fig. \ref{fig:cluster-maps}.

\section{Ablations}
\subsection{Gram Loss}
Gram loss \cite{simeoni2025dinov3} extends the mean squared error (MSE) loss, measuring the error between gram matrices of patch representation rather than the representations themselves. We compare gram loss to MSE for pretraining with the SA component. Table \ref{tab:gram-loss} shows that using the gram loss yields better downstream segmentation performance.

\begin{table}[hbt]
	\centering
	\caption{Ablation on gram loss using LatentMIM}\label{tab:gram-loss}
	\begin{tabular}{l *{4}{l}}
		\toprule
		& \multicolumn{4}{c}{\textbf{mIoU}} \\ \cmidrule(lr){2-5}
		\textbf{Model} & {\textbf{m-SA}} & {\textbf{Sen1Floods11}} & {\textbf{PASTIS}} & {\textbf{MADOS}} \\ 
		\midrule
		\rowcolor{gray!30}
        Gram loss & \res{\mathbf{23.39}}{0.02} & \res{\mathbf{78.82}}{0.12} & \res{\mathbf{22.23}}{0.04} & \res{\mathbf{43.00}}{0.84} \\
		
        MSE & \res{22.49}{0.07} & \res{78.52}{0.09} & \res{18.99}{0.04} & \res{38.37}{0.33} \\
		\bottomrule
	\end{tabular}
\end{table}

\subsection{High-resolution Representation Downsampling}
Due to the size difference stated in Eq. \ref{eq:scale} between the MR and HR images, each patch in the MR image corresponds to $s^2$ patches in the HR conterpart. Therefore, before applying the gram loss, we downsample the HR patch representations to match the size of the MR representations as stated in Sec. \ref{sec:method}. We test three downsampling methods: bilinear and bicubic interpolation, and linear projection. Table \ref{tab:dsample-methods} shows that bilinear downsampling yields better or comparable results despite being the most compute-efficient.

\begin{table}[hbt]
	\centering
	\caption{Comparison of downsampling methods in spatial affinity component with I-JEPA}\label{tab:dsample-methods}
	\begin{tabular}{l *{2}{l}}
		\toprule
		& \multicolumn{2}{c}{\textbf{mIoU}} \\ \cmidrule(lr){2-3}
		\textbf{Model} & {\textbf{m-SA}} & {\textbf{Sen1Floods11}}\\ 
		\midrule
		\rowcolor{gray!30}
        Bilinear          & \res{\mathbf{22.36}}{0.00} & \res{\mathbf{79.04}}{0.00} \\
		Bicubic           & \res{21.62}{0.01} & \res{78.28}{0.00} \\
		Linear projection & \res{21.50}{0.01} & \res{\mathbf{79.03}}{0.00} \\
		\bottomrule
	\end{tabular}
\end{table}

\subsection{Sampling strategy}
The spatial affinity component does not use all patches of its input image but rather samples a portion of the patches. Hypothesizing that contiguous patches carry spatial information better than random patches, we compare block sampling and LatentMIM's default random sampling. Table \ref{tab:sa-sampling} shows that using block sampling does not improve performance over using LatentMIM's default random sampling. 

\begin{table}[hbt]
	\centering
	\caption{Spatial affinity component sampling strategy ablations on LatentMIM}\label{tab:sa-sampling}
	\begin{tabular}{lll}
		\toprule
		& \multicolumn{2}{c}{\textbf{mIoU}} \\ \cmidrule(lr){2-3}
		\textbf{Model} & {\textbf{m-SA}} & {\textbf{Sen1Floods11}} \\ 
		\midrule
		  Default random & \res{23.39}{0.02} & \res{78.82}{0.12} \\
		  Block          & \res{21.43}{0.21} & \res{78.66}{0.15} \\
		\bottomrule
	\end{tabular}
\end{table}




\section{Discussions}
Our experiments show that integrating high-resolution satellite imagery data into mid-resolution pretraining using the spatial affinity component improves downstream semantic segmentation performance across diverse mid-resolution tasks over models pretrained with high- or mid-resolution imagery alone. This shows that effective methods of training with both HR and MR image data can exploit the advantages of either data to outperform models trained on only one of them. Given the scarcity of HR satellite imagery and thus the difficulty of acquiring good HR/MR image pair datasets, a promising direction for future work is to find methods to reduce reliance on good-quality image pairs.

\small
\bibliographystyle{IEEEtranN}
\bibliography{references, references_mendeley}

\end{document}